# BioREx: Improving Biomedical Relation Extraction by Leveraging Heterogeneous Datasets

Po-Ting Lai[a,†], Chih-Hsuan Wei [a,†], Ling Luo[b], Qingyu Chen[a], Zhiyong Lu [a,*]

[a] National Center for Biotechnology Information (NCBI), National Library of Medicine (NLM), National Institutes of Health (NIH), MD, 20894, Bethesda, USA

[b] School of Computer Science and Technology, Dalian University of Technology, 116024, Dalian, China

**Abstract**

Objective: Biomedical relation extraction (RE) is the task of automatically identifying and characterizing relations between biomedical concepts from free text. RE is a central task in biomedical natural language processing (NLP) research and plays a critical role in many downstream applications, such as literature-based discovery and knowledge graph construction. State-of-the-art methods were used primarily to train machine learning models on individual RE datasets, such as protein-protein interaction and chemical-induced disease relation. Manual dataset annotation, however, is highly expensive and time-consuming, as it requires domain knowledge. Existing RE datasets are usually domain-specific or small, which limits the development of generalized and high-performing RE models. Methods: In this work, we present a novel framework for systematically addressing the data heterogeneity of individual datasets and combining them into a large dataset. Based on the framework and dataset, we report on BioREx, a data-centric approach for extracting relations. Results and Conclusion: Our evaluation shows that BioREx achieves significantly higher performance than the benchmark system trained on the individual dataset, setting a new SOTA from 74.4% to 79.6% in F-1 measure on the recently released BioRED corpus. We further demonstrate that the combined dataset can improve performance for five different RE tasks. In addition, we show that on average BioREx compares favorably to current best-performing methods such as transfer learning and multi-task learning. Finally, we demonstrate BioREx's robustness and generalizability in two independent RE tasks not previously seen in training data: drug-drug N-ary combination and document-level gene-disease RE. The integrated dataset and optimized method have been packaged as a stand-alone tool available at https://github.com/ncbi/BioREx.

---

† The authors wish it to be known that, in their opinion, the first two authors should be regarded as joint first authors.
* Corresponding author. *E-mail address:* Zhiyong.Lu@nih.gov (Z. Lu).

## 1. Introduction

Biomedical literature is the primary knowledge source of novel biomedical research results. The literature contains rigorous statistical results and biological evidence, which are typically simplified to the relations among entities. With the rapid growth of biomedical literature, processing data volume becomes increasingly feasible by using relation extraction (RE) techniques. RE identifies the pairs of entities involved in the relations and assigns granular relation types. It effectively transforms unstructured text into structured knowledge. An example of this process is shown in Figure 1, which presents a portion of the abstract from PMID:31440061.

In this example, BRAF is a gene that normally regulates cell growth. However, a specific mutation in BRAF, known as V600E, can lead to uncontrolled cell growth, resulting in a tumor. Treating a tumor caused by the BRAF V600E mutation with a combination of the drugs dabrafenib and trametinib has been shown to inhibit or reduce tumor growth. The text essentially asserts the relations among the entities: BRAF V600E, tumor, trametinib, and dabrafenib. This exemplifies how RE can be utilized for performing literature-based discovery [3-6] and knowledge graph construction [6-8].

Various RE methods include co-occurrence [11, 12], rule-based approaches [14, 15], supervised methods [18, 19], and distant supervision [22, 23]. As the most straightforward method, co-occurrence usually achieves high recall at the expense of precision. The method involves collection of the pairs of entities that co-occur in single sentences or documents. Rule-based approaches, in contrast, can achieve higher precision by delineating predicates within relations. It is very difficult, however, to generate robust rules to handle all cases, which results in lower recall. In the past decade, many manually curated RE datasets [9, 10, 16, 24, 25] have been developed for public usage, and machine learning (ML) and deep learning (DL) methods have been widely employed [18, 19, 26-28] to solve RE tasks (e.g., chemical-disease [16, 29, 30], chemical-protein [9, 26-28], protein-protein [2, 13, 31], drug-drug [26-28, 32], and disease-gene [22, 29]).

Most RE datasets, however, contain only a single relation type, and these benchmark datasets are gathered from different annotation scopes. It is challenging and time-consuming to integrate different RE systems. For example, a protein drug in DrugProt [9] is categorized as a chemical, whereas AIMed [2] categorizes it as a gene product. Most systems were developed and validated on individual datasets, and this severely limited the development of generalized biomedical RE systems. In response, a new RE dataset, BioRED[33], was created, consisting of multiple relations (e.g., gene-disease) between some of the most important biomedical concepts (e.g., gene, disease, and chemical). The dataset was successfully used by the recent NIH LitCoin NLP Challenge[1]. Although BioRED is rich in relation types and has a broad annotation scope, it currently contains annotations of only 600 PubMed articles. Its portability and generalization abilities require further studies.

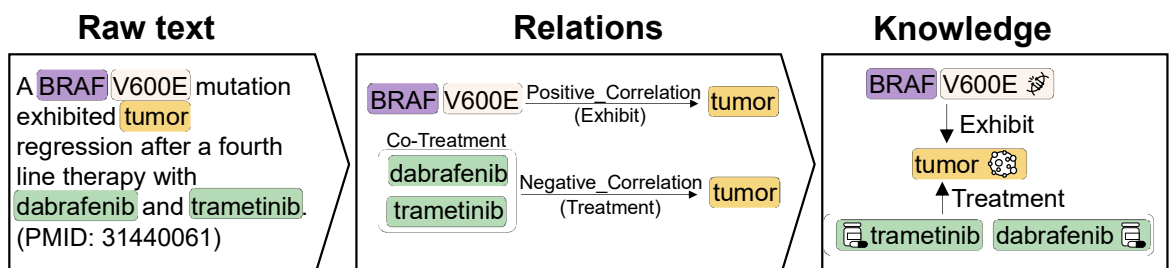

**Figure 1.** Example of the relations (middle) in the text (left) and the extracted knowledge (right).

There has been some research [34-36] on transfer learning (TL) and multi-task learning (MTL) for biomedical RE tasks. TL pretrains the language model on large datasets and fine-tunes the model in downstream task(s). As a result, downstream tasks benefit from the pre-trained language model (PLM), which has learned domain-specific neural networks, and can focus on tuning the task-specific neural parameters. In contrast, MTL involves many related tasks simultaneously and improves overall generalization, and MTL makes no distinction among the different tasks. Using these approaches, TL and MTL aggregate external data and eliminate the sparsity problem of small tasks.

This work proposes a method to improve the performance of all kinds of relation extraction tasks over state-of-the-art performance. Different from TL and MTL, a data-centric approach is proposed that harmonizes the annotation differences between data sources and integrates them to construct a rich dataset with sufficient quantity. Recognizing the nature of different annotation scopes in the existing datasets, we propose a harmonization framework for effectively improving the consistency of different annotations. The main contributions of our work can be summarized as follows:

(1) We propose a framework that can systematically integrate heterogeneous annotations into one large-scale dataset despite a diverse annotation scope and guidelines (e.g., relation annotations with or without entity spans).

(2) Our proposed approach, which is BioREx, results in a large training set for developing generalized RE systems. Using our combined dataset, higher performance is achieved compared with TL, MTL, and that of models trained on the individual datasets.

(3) Our results show that the pre-trained model built on the merged dataset is robust and generalizable through its applications to two new RE tasks that were not previously studied.

(4) Both the merged dataset and trained model are made freely available to the research community.

---

[1] https://ncats.nih.gov/funding/challenges/litcoin

## 2. Related works

### 2.1. Multi-task Learning for Relation Extraction

Multi-task learning (MTL) focuses on jointly learning different tasks. This is typically achieved through (1) a shared representation, which encodes different tasks into the same semantic space and (2) a sum of the losses from individual tasks, where the model's weights are updated using backpropagation against the combined losses obtained from each task in a batch.

In NLP, for instance, Zhao et al [37], Wiatrak et al [38], and Zhou et al. [39] successfully applied MTL to jointly learned Named Entity Recognition (NER) and entity linking tasks. Some studies have also explored MTL across NER and Relation Extraction (RE) tasks. Eberts et al. [40] proposed a transformer-based approach to jointly train NER and RE on the adverse drug effect (ADE) dataset [41], reporting promising results. However, it is important to note that even when these MTL studies are applied to different tasks, the annotations used still come from a single corpus.

Alternatively, MTL can be applied to a single NLP task using multiple corpora. Peng et al [36] demonstrated the effectiveness of this approach by learning jointly from eight different corpora spanning four different tasks and demonstrating significant improvements in all four tasks. However, they expressed concerns about the application of heterogeneous datasets to the task of relation extraction. As an example, the CDR corpus, which is a document-level chemical-disease relation extraction dataset, may not be suitable for the sentence-level drug-drug interaction [10], chemical-protein relation [42] tasks.

### 2.2. Document-level and N-ary Relation Extraction

Document-level RE identifies and classifies relations between multiple entity pairs within a single document [16, 22, 33]. In contrast to sentence-level RE [9, 10], which identifies and classifies relations for only one entity pair in each sentence, document-level RE requires the model to identify and classify relations for every entity pair in the document. Pairs can be related across multiple sentences. This task challenges the identification of relevant context for each entity pair. It has received attention in both the general and biomedical domains recently. Common approaches for document-level RE, like treating different pairs as individual instances, and applying deep learning models for classification[30]. To obtain relation labels, some studies aggregate the entity-level and document-level representations and feed them into a classifier [43]. Adaptive Transfer Learning with Local Context Pooling (ATLOP) [29] is an approach that leverages local context pooling to differentiate the same entity embedding across various entity pairs. It improves entity embedding by incorporating additional context relevant to the current entity pair. By transferring attention heads PLM to entities, ATLOP captures each entity's attention. Furthermore, it combines the attentions of both entities to identify the context that holds significance for both. This process enables ATLOP to enhance entity representation by incorporating relevant contextual information. ATLOP is evaluated on both general, DocRED dataset [44], and biomedical domains, BC5CDR and RENET [22] datasets. RENET is a large-scale distant supervision dataset that uses PubTator's annotations and hand-annotated disease-gene tuples in DisGeNet [45], with approximately 31,000 abstracts of disease-gene associations, and 1,000 manually re-annotated abstracts are used for evaluation.

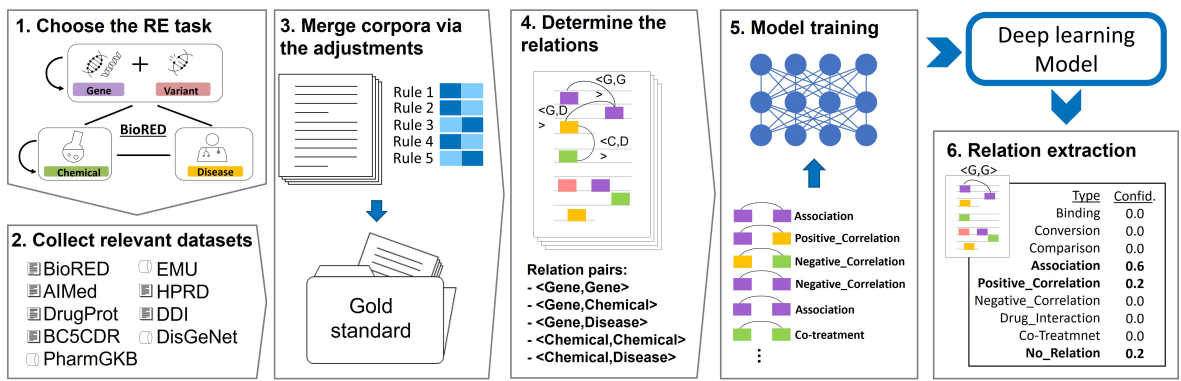

**Figure 2.** Workflow of our approach.

On the other hand, Tiktinsky et al. [46] developed a drug combination dataset for RE. It is intended to assist in the identification of effective combinations of therapies for diseases such as cancer, tuberculosis, malaria, and HIV. The set contains 1600 manually annotated abstracts mentioning between two and fifteen drugs. In the dataset, 840 abstracts have one or more positive drug combinations ranging from two to eleven drugs. There are 760 abstracts that either mention drugs that are not used in combination or discuss combinations of drugs that do not provide a positive result when used together. As a basis for further research and improvement, a PubMedBERT baseline model is developed to identify effective drug combinations. A co-occurring sentence was used to select drug subsets as instances for prediction, and greedy heuristics and unions were used to determine the final N-ary drug combination.

## 3. Material and methods

### 3.1. Workflow of BioREx

We propose a harmonization framework for unifying different datasets into one via addressing their heterogeneity. Figure 2 illustrates our workflow, which consists of four steps. The first step is to determine the target relation extraction task. As a case for this study, we chose BioRED, which focuses on the relations between genes (proteins), variants, diseases, and chemicals. As previously defined in BioRED, we continued to explore the eight relation types (positive correlation, negative correlation, association, bind, drug-interaction, co-treatment, comparison, and conversion). Further, entities with interchangeable names are combined (e.g., gene products are grouped together with genes, drugs are grouped with chemicals, phenotypes are grouped with diseases). Moreover, we simplified the BioRED task by merging variants and genes into a single concept and their relationships accordingly, as the mutated genes are much more frequently searched in PubMed as compared to the variants.

The second step is to collect datasets relevant to user research interests and benchmarking tasks. There are eight relevant datasets selected for the BioRED task in total, including corpora and repositories. Text corpora are collections of texts with manually curated entities and the relations between them. The relations are typically at the mention level. In contrast, a repository contains a list of relations at the document level. For example, in the CTD repository [47], relation triples, such as <entity1, entity2, PMID>, consist of gene, disease, and chemical entities with PMID, but no mention level annotations are provided. In addition, full text versions can be used to curate relations. In the third step, we systematically adjust the datasets to a uniform format by using our procedure, enabling them to be merged with BioRED. In the fourth step, we formulate the RE task as a classification problem for each relation pair. The fifth step involves training a deep learning classifier using a state-of-the-art BERT-based pre-trained language model (PLM). In the last step, the well-trained model can be applied to the new articles for relation extraction.

## 3.2. Data collection

We summarized most of the existing RE datasets in the supplementary material (Table S1). In our observation, it is more difficult for some to be integrated into a large RE dataset due to several issues:

(1) **Not fully annotated:** As shown in Table S1, some datasets were built by the automatically extracted entities and relations. Certain studies refined those automatic results but did not curate the entities or pairs from scratch. Lower annotation consistency or a lack of annotation guidelines could result in negative effects if we integrated these datasets with other datasets.

(2) **Specific scope:** Some datasets focus on specific subsets of granular entity and relation types and do not annotate all entities and relations. For example, ADE [41] annotates the relations among drug, dosage, and adverse effect but only partially matches the scope of chemical, disease, and their relations in BioRED. To avoid the mismatch of scopes that may result in an inter-dataset annotation inconsistency, we do not use certain datasets.

(3) **No entity spans:** Many repositories and a few datasets do not include relation entity spans. Based on our procedure, the automatic annotation of PubTator [48] should be applied to complement the entity spans. PubTator's annotation, however, is not entirely accurate. Specifically, ML and DL models may be confused by a missed or incorrect annotation.

To enhance the consistency of the integrated dataset, we first chose the datasets (i.e., AIMed [2], DrugProt [9], DDI [10], HPRD50 [13], and BC5CDR [16]) that are relatively free from the above issues due to clear guidelines and well-defined entities. Due to lack of a publicly available dataset for the <G, D>

relation, we collected EMU [17] in the dataset pool and DisGeNet [45] repository. In addition, we include PharmGKB [20] to support the <G, C> entity pairs that still have room for improvement.

### 3.3. Data adjustment procedure

Each dataset is characterized by its annotation scope and guidelines. For example, a relation pair annotated within the same sentence is a common occurrence; however, only a few datasets allow for cross-sentence relations. In addition, some datasets map the entities to specific concepts in controlled vocabularies (e.g., MESH [49]), while others do not. To integrate datasets into one dataset, we identify the characteristics of the datasets and define five categories of characteristic diversity. Five solutions are provided to adjust the different datasets as shown in Table 1. The details of the adjustment solution for the eight datasets are illustrated in Table 2.

(a) **With entity span:** In some datasets, such as repositories, the spans of entities are not provided. However, these spans are essential for training the RE model. To address the gap, we used both the dictionary match approach and the automatic annotation by PubTator [48] to map the corresponding spans of those entities into the dataset entities. As some of the entities cannot be found in the automatic annotations, we retained only those relations with spans found in the text. For example, the annotation from repositories, such as PharmGKB, is curated by reading full text versions. In that case, we retained only the relations with the associated spans in the same abstract sentence accordingly.

(b) **Document- or sentence-level:** This category distinguishes datasets by the context range of the association evidence. Relation datasets can be roughly divided into two levels, document- and sentence-levels, according to their annotations. (1) Document-level: The relation entities are allowed to appear across sentences, and their spans need to be given in the sentences. (2) Sentence-level: This level requires the entity pair and its evidence to co-occur in a single sentence. Typically, such datasets do not map entity spans to concept identifiers. Depending on the context range, which can be at the sentence-level or document-level, we delimit the input text to the corresponding sentence (or the document) as required. Because repository annotations are more reliable due to manual curation, we considered them sentence-level annotations.

(c) **Negative or unannotated relations:** Some datasets focus on only a specific scope of a relation pair. For instance, BC5CDR [16] annotates the chemical-induced disease relation but not the treatment. These unannotated pairs are not necessarily negative (i.e., they don't definitively lack the relation), they're simply unannotated, and thus it would be incorrect to mark all of them as negatives. To more effectively utilize annotations in this case, we assigned a specific relation type to the pairs that were not curated (e.g., "None-BC5CDR" for BC5CDR). Certainly, if the dataset annotates all the relations in the target entity pair, the specific relation type is not required for those unlabeled pairs. Besides, some datasets do not annotate all the relations in the text, such as the repositories (e.g., PharmGKB). In this case, we do not use those unannotated pairs as negative cases.

**Table 1.** Characteristic diversity and adjustment solutions.

| Characteristic diversity | Cases of the diversity | | Adjustment solutions |
|---|---|---|---|
| a. With entity spans | Yes | | a1. No change is needed. |
| | No | Relation is from abstract only. | a2. Use the automatic annotation and retain the relations while both entities are in the abstract. |
| | | Relation may be from full text (e.g., repositories). | a3. Use the automatic annotation and retain the relations while the two entities are in the same sentence. |
| b. Document- or sentence-level | Document-level | | b1. Use the whole abstract as the input. |
| | Sentence-level | | b2. Use the co-occurring sentence as the input. |
| c. Negative or unannotated relation pairs | Specific scope of the relation pair | | c1. Generate an internal negative class (e.g., "None-BC5CDR") for those pairs that were not curated. |
| | Consistent scope with target set | Complete annotation | c2. Train on those pairs that were not curated as the negative cases. |
| | | Incomplete annotation | c3. Do not train on those pairs that were not curated. |
| d. Granular relation type defined (Y/N) | Yes | | d1. Merge the granular types to Positive, Negative, and Association or other relation types. |
| | No | | d2. All pairs are deemed as Association. |
| e. Entity definition to the target dataset | Different | | e1. Assign an internal entity type (e.g., "DrugProt-Chem") to the entities. |
| | The same or similar | | e2. Use the same entity type tags as the target dataset (i.e., BioRED). |

(d) **Granular relation type defined (Y/N):** Some datasets focus on specific concept pairs (e.g., DrugProt focuses on the chemical-gene pair), with multiple granular relation types (e.g., inhibition). The adjustment solution to utilizing those annotations is to manually map the granular relation types to the corresponding relation types in BioRED. For example, DrugProt relation types, such as AGONIST-INHIBITOR, ANTAGONIST, INDIRECT-DOWNREGULATOR, and INHIBITOR, are converted into negative correlations, whereas ACTIVATOR, AGONIST, AGONIST-ACTIVATOR, and INDIRECT-UPREGULATOR are classified as positive correlations. If granular types cannot be mapped to any type in BioRED, the relations will be considered "Association."

**Table 2.** Corpora collected for training and evaluation. SEN: sentence level. DOC: document-level. No span: not entity span is provided. D = Disease, G = Gene/Protein, and C = Chemical/Drug.

| Corpus | # Abstracts | # Relation | Relation pair | SEN/DOC Level | Conditional category | | | | |
|---|---|---|---|---|---|---|---|---|---|
| | | | | | a | b | c | d | e |
| BioRED [1] | 600 | 6,503 | 8 pairs | DOC | a1 | b1 | c2 | d1 | e2 |
| AIMed [2] | 225 | 847 | <G,G> | SEN | a1 | b2 | c1 | d2 | e2 |
| DrugProt [9] | 4,250 | 21,035 | <G,C> | SEN | a1 | b2 | c1 | d1 | e1 |
| DDI [10] | 905 | 4,965 | <C,C> | SEN | a1 | b2 | c1 | d2 | e2 |
| HPRD50 [13] | 50 | 138 | <G,G> | SEN | a1 | b2 | c1 | d2 | e2 |
| BC5CDR [16] | 1,500 | 3,106 | <C,D> | DOC | a1 | b1 | c1 | d1 | e1 |
| EMU [17] | 89 | 115 | <G,D> | No span | a2 | b2 | c3 | d2 | e2 |
| PharmGKB [20] | 823 | 949 | <G,C> | No span | a3 | b2 | c3 | d2 | e2 |
| DisGeNet [21] | 2,803 | 859 | <G,D> | No span | a3 | b2 | c3 | d2 | e2 |

(e) **Entity definition of the target dataset:** Different datasets may define the same concept differently. For example, a protein drug in DrugProt is categorized as a drug, whereas BioRED categorizes it as a gene product. For these chemicals in DrugProt, we assigned a specific entity type (i.e., "DrugProt-Chem" instead of "Chemical"). A dataset can be merged without changing the types of entities if the entities match the BioRED.

We adjusted each dataset according to its own annotation characteristics. The selected datasets and the corresponding adjustments are shown in Table 2. For example, AIMed is a sentence-level dataset of protein-protein interaction (<G,G>). All of the text spans of the mentioned proteins in the sentence are annotated even though some are not asserted in a relation. Further, no granular type was annotated in AIMed. Based on these characteristics, we modified the AIMed accordingly: (a1) AIMed provides entities involved in relations; thus, no automatic entity annotation is needed. (b2) AIMed is a sentence-level dataset. Hence, we use the sentence with the co-occurring relation entities as the input. (c1) We generated a specific relation type, "None-AIMed," for those pairs that were not annotated in AIMed. (d2) Because no granular type was annotated to the relations in AIMed, we simply set all the relation pairs as "Association." (e2) The concept definition of the AIMed is the same as the "Gene" concept in BioRED, which we use consistently.

## 3.4. Deep learning model for relation extraction

In this section, we first introduce how we formulate the RE problem. We then present our deep learning method.

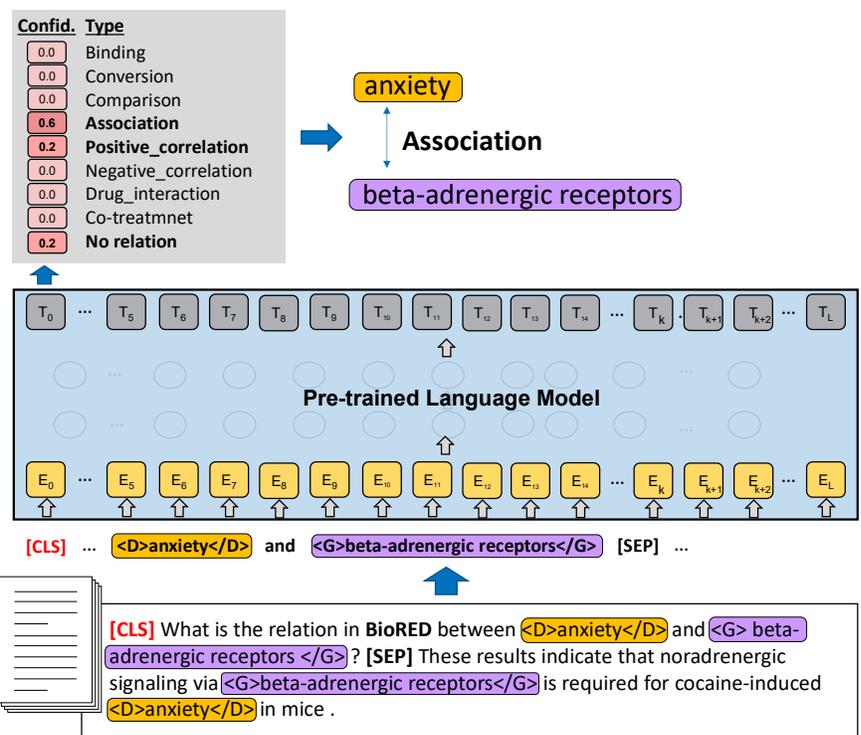

**Figure 3.** Architecture of the BioREx model on a relation extraction task.

**Problem formulation:** We defined a relation candidate instance that contains a pair of biomedical entities and their context of co-occurrence. A few datasets, such as BioRED and BC5CDR, annotate relations at the document level. Such datasets annotate all spans of the same entity, which may appear multiple times. In addition, if an entity span contains two or more entities (e.g., "breast or ovarian cancer" contains both MESH:D001943 and MESH:D010051), the relation with this span must be expanded to two (or more) instances. Each instance requires that two entities belong to unique IDs individually. We treat RE as text classification, which aims to classify the instance into a pre-defined RE relation type or no relation ("None").

**Deep learning model:** The architecture of the model of BioREx is illustrated in Figure 3. For each instance, two boundary tags are inserted at the beginning and the end of the entities (e.g., <D> and </D> for diseases). We also added those tags to the vocabulary of the PLM to ensure that the tags are not separated into multiple tokens. In addition, we further constructed a prompt question to provide contextual guidance to the model for its corresponding pair and the RE task. The prompt question, likes "[CLS] What is the relation in [Corpus] between <entity1> and <entity2>?", is appended at the beginning of the input text to emphasize the two entities in a pair (<entity1> and <entity2>) and the RE task ([Corpus]). [Corpus] is the specific task name for various tasks, such as "DrugProt." "BioRED" is the default of the [Corpus], and we used it for the BioRED task.

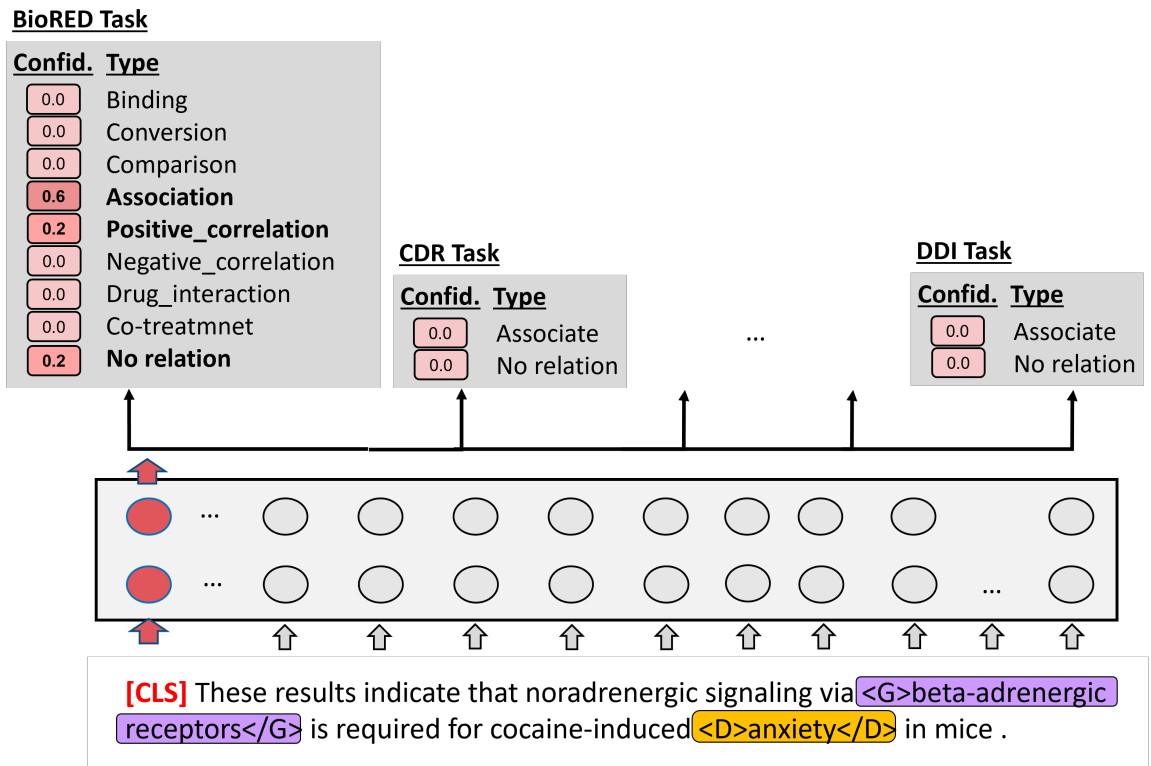

**Figure 4.** An example of multi-task learning (MTL) approach.

The confidence scores of the classification output depend on the state of the PLM's [CLS] token. In order to obtain the confidence scores for each class, the fixed-length representation of the [CLS] token is fed into a linear layer followed by a softmax activation. A prediction for the text sequence is based on the class with the highest probability. For our implementation, we chose PubMedBERT [28] as our default PLM. The output of the model provides the confidence of the correlation types (or no relation). The type with the highest confidence becomes the predicted output.

### 3.5. Approaches for comparison

To evaluate BioREx's effectiveness, we compare BioREx with two other deep learning configurations. (1) Transfer learning (TL): We trained the BERT model on the external datasets and then fine-tuned it on the training set of the evaluated task. (2) Multi-task learning (MTL): We consider different RE datasets as individual tasks using the multi-task learning approach [36] as depicted in Figure 4. In MTL, the distinguishing of different tasks does not rely on prompts. Instead, all datasets are combined and instances are randomly arranged. In this approach, the [CLS] tag is used as the representation fed into all tasks. In this approach, the [CLS] tag is used as the representation fed into all tasks. The [CLS] tag serves as a shared representation across the different tasks where each task has its own task-specific output layers. This enables the model to capture and leverage the underlying information from each task without task-specific prompts. By using this uniform representation, MTL addresses multiple tasks simultaneously. Furthermore, we evaluated the portability of our model, which was trained on BioRED along with all eight external datasets (BioRED+8 datasets). As shown in Figure 5, we fine-tuned the BERT model on the adjusted BioREx training set using the architecture described in section 3.4. Subsequently, the fine-tuned BERT model was adapted to Tikinsky et al.'s and ATLOP's source code implementations for drug combination N-ary and RENET tasks, respectively. This experiment is crucial to validate the BioREx's adaptability to different RE system architectures.

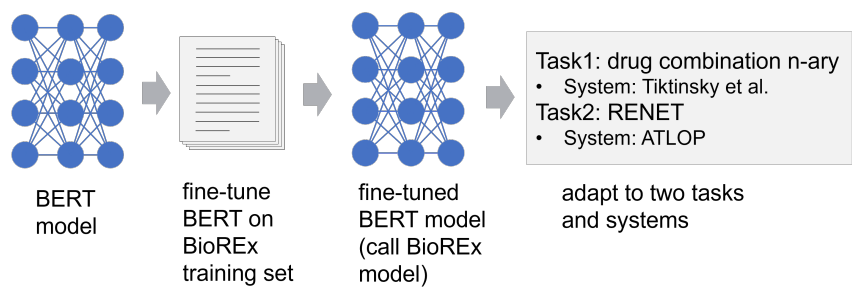

**Figure 5.** Evaluate the BioREx model using other tasks and open source systems.

## 4. Results

To demonstrate the advantage of our data-centric method for improving RE tasks, we design three experiments: (1) To demonstrate that BioREx can be used to improve the performance on a BioRED task, we train the model by using different training set combinations and evaluate the model on the BioRED test set. (2) To show the generalizability of BioREx on other datasets, we use BioRED as an external dataset for five different RE tasks. (3) To enhance the portability of our work, we release the language model (LM), which was trained on BioRED plus all external datasets (BioRED+8 datasets). The experiment evaluates the robustness and generalizability of the BioREx model by using two different tasks not previously seen in training data: N-ary [46] and RENET [22].

### 4.1. Performance on the BioRED task

In this experiment, we evaluate how external datasets improve BioRED performance. As seen in Table 3, the PubMedBERT model trained on the BioRED training set is set as the baseline (Original). We evaluated the performance on the BioRED test set by leveraging external datasets via the three configurations (i.e., TL, MTL, and BioREx).

To evaluate the effects of external datasets, we developed the models using the BioRED training set and the external datasets individually. The three datasets (i.e., DisGeNet, PharmGKB, and EMU) cannot be integrated into the TL and MTL training sets due to the lack of entity spans. We conduct a pairwise *t*-test to determine whether there is a significant difference in the mean F-score of the two models. As we expected, TL, MTL, and BioREx outperform the baseline with $p$-values of 0.029, 0.002, and 0.001, respectively. In addition, the model trained on the adjusted datasets (BioREx) performed better than did TL and MTL for most of the relation pairs, except for MTL, which is slightly higher than BioREx in BioRED+DDI and BioRED+AIMed.

**Table 3.** Performance on the BioRED task using external datasets. All numbers are F-scores. *$p < 0.05$ (pairwise $t$-test in mean F-score compared with baseline).

| Training set | Entity pair | Performance on BioRED test set | | | |
|---|---|---|---|---|---|
| | | Original | TL | MTL | BioREx |
| BioRED+BC5CDR [16] | <C,D> | 76.5 | 78.4 | 78.2 | **79.3** |
| BioRED+DDI [10] | <C,C> | 78.0 | 77.4 | **78.6** | 78.3 |
| BioRED+AIMed [2] | <G,G> | 78.1 | 75.7 | **82.0** | 81.5 |
| BioRED+HPRD50 [13] | | | 79.9 | 78.1 | **81.9** |
| BioRED+DrugProt [9] | <G,C> | 78.3 | 80.7 | 82.8 | **85.8** |
| BioRED+PharmGKB [20] | | | - | - | **79.7** |
| BioRED+DisGeNet [45] | <G,D> | 67.2 | - | - | **71.2** |
| BioRED+EMU [17] | | | - | - | **69.7** |
| BioRED+8 datasets | All | 74.4 | 77.6* | 78.0* | **79.6*** |

We analyzed Table 3 results, and carefully examined some cases where MTL produced true positives (TP) but BioREx generated either false negatives (FN) or false positives (FP) within the BioRED+DDI and BioRED+AIMed datasets. Further, we observed that out of the randomly selected 15 cases, 11 had more than one coinciding sentence in the abstract. However, their relational statements were not explicitly defined. For instance, in PMID:23069675, *"Our results indicate a regulatory role of NP1 in Bad/Bax - dependent mitochondrial release of Cyt C and caspase-3 activation."* The relation between the gene-gene relation pairs <Bad, Cyt C, Association>, <Bad, caspase-3, Association>, <Bax, Cyt C, Association>, and <Bax, caspase-3, Association><Bad, caspase-3, Association> are not clearly stated. Similarly, the chemical pair <tacrolimus, everolimus, Conversion> from PMID:18503483 *"Recovery of tacrolimus-associated brachial neuritis after conversion to everolimus in a pediatric renal transplant recipient--case report and review of the literature"* is only implicitly connected since the conversion mechanism is not explicitly explained. It appears that MTL has a slight edge over BioREx in identifying these implicit relations.

Additionally, we also examined the instances where BioREx generated TP but MTL produced FN/FP within the BioRED+HPRD50 and BioRED+DrugProt datasets. We found that the majority of these instances were low-frequency pairs, where at least one entity appears only once in the abstract. BioREx identified these pairs correctly, while MTL did not. Consequently, we evaluated the identification of low-frequency pairs across the entire BioRED test set. The results revealed that Original, TL, MTL, and BioREx had F-scores of 46.0%, 55.7%, 56.2%, and 61.5%, respectively. Since BioREx utilizes the entire model for various tasks, including sentence-level tasks that tag an entity only once per instance, it significantly outperforms the other methods.

Document-level annotations (e.g., BC5CDR) and sentence-level annotations (e.g., DDI) serve distinct goals and offer different advantages. Sentence-level annotations provide precise context, including the locations of entity pairs and supporting evidence. On the other hand, document-level annotations do not offer explicit relation sentences between pairs but provide rich contextual information. By merging datasets from both document- and sentence-level annotations, our model becomes capable of handling different types of contexts, making it more versatile. Ours has improved the DDI performance by 2.7% compared with the baseline, but with a $p$-value of 0.11, which is not statistically significant.

The performance of BioREx on chemical-disease <C,D> relations increased from 76.5% to 79.3% after we merged the BC5CDR dataset with BioRED, which also shows better performance than the models trained by TL (77.6%) and MTL (78.0%). The performance of the gene-chemical <G,C> relation also was improved when we merged the adjusted datasets of DrugProt or PharmGKB within BioRED. Further, PharmGKB is a repository with manual annotation from full text, which means some of the annotations exist in the full text but not in the abstract. In addition, not all the relations in an article are recorded in PharmGKB.

Finally, we trained the models using the training set of BioRED plus all external datasets (BioRED+8 datasets) in the three configurations. BioREx achieved the highest performance (79.6% in F-score), which is above 2–3% of other methods. Compared to TL and MTL, BioREx is higher than theirs by F-scores of 2.0% and 1.6%, respectively. Further statistical results, however, show that improvement did not pass the significant statistical test with $p$-values (TL vs. BioREx: 0.4; MTL vs. BioREx: 0.89). By our observations, the harmonization procedure reduced the heterogeneity between datasets and is able to integrate those into a large dataset.

We have conducted an analysis of the errors made by the TL models trained on the BioRED+DDI and BioRED+AIMed datasets. These errors specifically pertain to relation pairs that were correctly recognized by the original model. Upon reviewing the missed relation pairs, we observed that the entities involved in these pairs often co-occur within the same sentence. However, the evidence statements supporting these relationships are not explicitly stated. As we observed, corpora have different curation criteria, such as the DDI dataset not annotating co-treatment relations between drugs. Without using the harmonization procedure to reconcile the discrepancy of the annotations, the transfer learning process may not yield a reasonable result. We are actively working to address these issues and improve the performance of our models.

**Table 4.** Performance of the RE model on the test set of individual datasets. All numbers are performance (F-score) of the test set in individual datasets. *$p < 0.05$ (pairwise *t*-test in mean F-score compared with baseline).

| Test set | relation | Target dataset | Target dataset + BioRED | | | SOTA |
|---|---|---|---|---|---|---|
| | | | TL | MTL | BioREx | |
| BioRED | All | 74.4 | 77.6 | 78.0 | **79.6** | 74.4[33] |
| BC5CDR | \<D,C> | 69.1 | 68.4 | 69.3 | **69.9** | 69.4[29] |
| DDI | \<C,C> | 85.0 | 85.6 | 85.9 | **87.3** | - |
| DrugProt | \<G,C> | **80.9** | 80.4 | 80.3 | 80.8 | - |
| AIMed | \<G,G> | 82.4 | 84.1 | 84.2 | **84.9** | 83.4[50] |
| HPRD50 | \<G,G> | 83.0 | 84.2 | 85.6 | **86.2** | 84.5[50] |
| Average | All | 79.1 | 80.1 | 80.6* | 81.4* | - |

## 4.2. Performance on individual datasets by leveraging BioRED dataset

In the previous experiment, we demonstrate that the proposed procedure leverages external datasets to improve performance on the BioRED task. We are interested in determining whether BioREx can be applied to other tasks. Here, we focus on the five benchmark datasets that provide the entity spans in the texts. We used BioRED as an external dataset to support individual tasks. For each task, a model was first trained and evaluated on the target dataset (e.g., DDI). Based on the three configurations, we developed three models and tested them on the test sets for each dataset. In addition, we compared the models with SOTA approaches [29, 33, 50]. Our configurations are compared to the BioRED benchmark result. In BC5CDR, we compare the configurations with ATLOP [29], which does not use any additional post-processing. As a result of our procedure, DrugProt and DDI relation types were combined into BioRED relation types, which makes us unable to compare our configurations with SOTA [9, 51] directly. Further, AIMed and HPRD50 do not have formal partitions for training and testing. Therefore, we follow most studies by using 10-fold cross-validation to evaluate different configurations (Target dataset, TL, MTL, and BioREx). Note that some of the datasets (e.g., PharmGKB) do not annotate all the relations in the articles, which means that we were not able to locate those entity pairs that are not in the relations. Therefore, we omitted those datasets in this experiment.

As an example of the evaluation of the AIMed dataset, the model training on its training set obtained an F-score of 82.4%. Once the two datasets are merged with adjustments, the model presents a reliable performance on both test sets, which are both higher than the baselines (82.4% to 84.9 % in AIMed test set (Table 4) and 78.1% to 81.5% in BioRED (Table 3)). The three other models do not achieve better than BioREx. Despite all models' having improvements over the baseline, statistical significance is not achieved; the p-values are 0.37, 0.15, and 0.06 for TL, MTL, and BioREx, respectively. Both MTL and BioREx outperform the baseline with p-values of 0.045 and 0.044, respectively.

**Table 5.** Evaluation of N-ary based on different PLMs. PubMedBERT is the default PLM model. State-of-the-art (SOTA) performance has been reported in the N-ary study.

| Evaluation | PLM | Performance (F-score) | |
| | | Exact match | Partial match |
| --- | --- | --- | --- |
| Positive Combination | PubMedBERT (SOTA) | 65.2 | **72.4** |
| | BioRED | 64.8 | 71.0 |
| | BioREx | **66.2** | 71.4 |
| Any Combination | PubMedBERT (SOTA) | 70.8 | 78.7 |
| | BioRED | 68.0 | 75.2 |
| | BioREx | **75.8** | **81.8** |

**Table 6.** Evaluation on RENET based on different PLMs. SciBERT presents the SOTA performance reported in the RENET study.

| Method | PLM | Precision | Recall | F-score |
| --- | --- | --- | --- | --- |
| ATLOP | SciBERT (SOTA) | 81.5 | 86.4 | 83.9 |
| ATLOP | BioRED | 81.6 | 85.4 | 83.5 |
| ATLOP | BioREx | **81.8** | **86.8** | **84.2** |

As we expected, the performances of TL and MTL are both improved; but they are still 1.3% and 0.8% lower in F-scores on average compared to our approach. Table 4 shows how important consistency between datasets is to performing multiple tasks.

4.3. Evaluating the robustness and generalizability of pre-trained language models

To demonstrate the robustness and generalizability of our pre-trained models, we chose two RE tasks, N-ary [52] and RENET [22], that have not been studied in BioRED or the other eight datasets. To demonstrate the portability of our PLMs, we used Tiktinsky et al.'s [52] and ATLOP [29] state-of-the-art implementations for the N-ary and RENET datasets, respectively.

Drug combination relations in the N-ary dataset: We followed the experiment in the N-ary study [52], which differentiated the relations into three categories: Positive Combination (POS_COMB), Non-positive Combination (OTHER_COMB), and Not a Combination (NO_COMB). Using the Positive Combination evaluation metric, the OTHER_COMB is grouped with NO_COMB. Using the Any Combination evaluation metric, the OTHER_COMB is grouped with POS_COMB. The drug combination relations may contain two or more drugs. The exact match requires all relation pairs to be extracted. Other than the exact match, the performance of the partial match determined only a subset of relation pairs in the combination.

As shown in Table 5, the state-of-the-art (SOTA) result was achieved by the proposed method in [52], using PubMedBERT as the PLM. Therefore, we retrained the model, using their open-source implementation [52] and replaced PLM with BioRED and BioREx pre-trained models. As can be seen, the BioREx model outperforms the baseline (PubMedBERT) by 1% and 5% in the Positive and Any Combinations, respectively.

Gene-disease association in the RENET dataset: Other than the relations and no-relations of the disease-gene pairs in the RENET dataset, RENET dataset categorized the semantically ambiguous pairs into ambiguous relations. We applied the proposed method, which uses ATLOP [29] with the default PLM of SciBERT [53]. We further applied ATLOP, using our pre-trained models (BioRED and BioREx PLMs). As shown in Table 6, the configuration of ATLOP trained on the BioREx PLM presents the highest performance.

## 5. Discussion

To demonstrate the effectiveness of our method on smaller datasets, we conducted an evaluation on the models trained using varying sizes of BioRED dataset, which is included in the supplementary materials (Experiment A). Our findings indicate that our approach has the capability to leverage the external datasets to build dependable training data with minimal annotation requirements.

While BioREx has exhibited encouraging results for relation extraction, its performance is far from perfect. To gain a deeper understanding of remaining challenges, we examined the results of the BioRED test set, with particular emphasis on the findings of the most effective model (BioREx). Through a random selection process, we analyzed 40 instances of errors and have subsequently presented the distinguishing features of these cases. It is worth noting that some errors may demonstrate more than one attribute.

The primary category of error (83%) is associated with low-frequency entities featured in the abstract. Most of these errors involve at least one entity that has been mentioned only once in the abstract. For instance, in the abstract of PMID:15485686, it states, *"These findings suggest that the Na(v)1.5/V1763M channel dysfunction and possible neighboring mutants contribute to a persistent inward current due to altered inactivation kinetics and clinically congenital LQTS with perinatal onset of arrhythmias that responded to lidocaine and mexiletine."* Here, there is a gene-disease association <Na(v)1.5, arrhythmias, Association> in the last sentence of the abstract. However, the association description is not explicit in the context, and *"arrhythmias"* appears only once in the abstract. As a result, BioREx fails to identify this pair, leading to a false negative. Among the errors, 63% have been classified as false negatives, while 20% are identified as false positives. These error cases frequently do not pertain to the key relations elaborated in the abstract, Consequently, BioREx is unable to extract the necessary information to accurately classify these pairs, leading to their classification as no relation instances.

The second type of error (30%) pertains to the absence of sentences where entities co-occur. In BioRED, there are approximately 15% of relations that lack co-occurring sentences, rendering them more challenging to extract. Due to the lack of supporting co-occurrence evidence, BioREx is inclined to predict these cases as no relational instances. Aomng these errors, 22.5% are classified as false negatives, while 7.5% are predicted as false positives. For example, in PMID:18503483, it states, *"MRI demonstrated hyperintense T2 signals in the cervical cord and right brachial plexus roots indicative of both myelitis and right brachial plexitis. Symptoms persisted for three months despite TAC dose reduction, administration of IVIG and four doses of methylprednisolone pulse therapy."* In this case, the disease *"brachial plexitis"* and the chemical *"methylprednisolone"* do not have any co-occurring sentence in the abstract. Based on the context, BioREx predicts an "Association" between them, but since the relation is not annotated in BioRED, it leads to a false positive case.

The last category of errors (13%) encompasses a set of challenging error types that defy simple categorization. These errors involve co-occurring entities that are mentioned multiple times in the text. For example, PMID:18457324 entitled *"Genetic polymorphisms in the carbonyl reductase 3 (CBR3) and the NAD(P)H:quinone oxidoreductase 1 (NQO1) genes in patients who developed anthracycline-related congestive heart failure after childhood cancer,"* the paper primarily discusses the gene expression of congestive heart failure (CHF) patients who had childhood cancer. Although the term *"cancer"* occurs multiple times in the abstract, there is no association between *"cancer"* and those genes related to CHF. However, BioREx incorrectly predicts them as positive instances. Consequently, 7.5% of these errors are erroneously classified as positive instances, while 5.5% are inaccurately classified as negative instances.

We further observed the errors of the results and realized that most remaining errors are due to two main challenges. First, a significant *p*-value or genomic evidence may be required to demonstrate the relation. Thus, the genetic and statistical evidence of the relation frequently comes from multiple sentences, which causes the largest portion of errors. Second, a difficulty arises when some of the relation statements (e.g., previous studies) can be overturned by following sentences (e.g., the authors present an interesting finding that conflicts with the initial assumption of the related study).

## 6. Conclusions

Using BioREx to integrate diverse datasets within different relation topics (e.g., gene-gene and chemical-disease relations) and criteria (e.g., sentence- and document-levels), the new model performs better than do the models trained on the individual dataset. Many available bioconcept relation resources have emerged during the past decade. This study proposed a procedure to improve the consistencies of the heterogeneous datasets and further optimize the performance. As in the case study on the N-ary and RENET datasets, we obtained a promising result, which demonstrated the possible usage to other RE tasks. To support other tasks, we have released the adjusted union dataset and the well-trained model for stand-alone usage. In the future, we will use BioREx on other repositories (e.g., CTD [47]) and further apply it to large-scale data, such as entire PubMed abstracts and PMC full-text versions.

**Acknowledgements**

This research was supported by the NIH Intramural Research Program, National Library of Medicine. It was also supported by the National Library of Medicine of the National Institutes of Health under award number 1K99LM014024 (Q. Chen) and the Central Universities [DUT23RC(3)014 to L.L.].

**References**

[1] Luo L, Lai P-T, Wei C-H, Arighi CN, Lu Z. BioRED: A Rich Biomedical Relation Extraction Dataset. (in press) Briefings in Bioinformatics. 2022.

[2] Bunescu R, Ge R, Kate RJ, Marcotte EM, Mooney RJ, Ramani AK, et al. Comparative experiments on learning information extractors for proteins and their interactions. Artificial Intelligence in Medicine. 2005;33:139-55.

[3] Gopalakrishnan V, Jha K, Jin W, Zhang A. A survey on literature based discovery approaches in biomedical domain. Journal of Biomedical Informatics. 2019;93:103141.

[4] Pyysalo S, Baker S, Ali I, Haselwimmer S, Shah T, Young A, et al. LION LBD: a literature-based discovery system for cancer biology. Bioinformatics. 2018;35:1553-61.

[5] Pilehvar MT, Bernard A, Smedley D, Collier N. PheneBank: a literature-based database of phenotypes. Bioinformatics. 2021;38:1179-80.

[6] Schutte D, Vasilakes J, Bompelli A, Zhou Y, Fiszman M, Xu H, et al. Discovering novel drug-supplement interactions using SuppKG generated from the biomedical literature. Journal of Biomedical Informatics. 2022;131:104120.

[7] Liu Y, Elsworth B, Erola P, Haberland V, Hemani G, Lyon M, et al. EpiGraphDB: a database and data mining platform for health data science. Bioinformatics. 2021;37:1304-11.

[8] Malec SA, Wei P, Bernstam EV, Boyce RD, Cohen T. Using computable knowledge mined from the literature to elucidate confounders for EHR-based pharmacovigilance. Journal of Biomedical Informatics. 2021;117:103719.

[9] Miranda A, Mehryary F, Luoma J, Pyysalo S, Valencia A, Krallinger M. Overview of DrugProt BioCreative VII track: quality evaluation and large scale text mining of drug-gene/protein relations. Proceedings of the seventh BioCreative challenge evaluation workshop2021.

[10] Herrero-Zazo M, Segura-Bedmar I, Martínez P, Declerck T. The DDI corpus: An annotated corpus with pharmacological substances and drug–drug interactions. Journal of Biomedical Informatics. 2013;46:914-20.

[11] Stapley BJ, Benoit G. Biobibliometrics: information retrieval and visualization from co-occurrences of gene names in Medline abstracts. Biocomputing 2000: World Scientific; 1999. p. 529-40.

[12] Jenssen T-K, Lægreid A, Komorowski J, Hovig E. A literature network of human genes for high-throughput analysis of gene expression. Nature Genetics. 2001;28:21-8.

[13] Fundel K, Küffner R, Zimmer R. RelEx—Relation extraction using dependency parse trees. Bioinformatics. 2007;23:365-71.

[14] Li Q, Wang X, Zhang Y, Ling F, Wu CH, Han J. Pattern discovery for wide-window open information extraction in biomedical literature. 2018 IEEE International Conference on Bioinformatics and Biomedicine (BIBM): IEEE; 2018. p. 420-7.

[15] Huang M, Zhu X, Hao Y, Payan DG, Qu K, Li M. Discovering patterns to extract protein–protein

interactions from full texts. Bioinformatics. 2004;20:3604-12.

[16] Wei C-H, Peng Y, Leaman R, Davis AP, Mattingly CJ, Li J, et al. Assessing the state of the art in biomedical relation extraction: overview of the BioCreative V chemical-disease relation (CDR) task. Database: The Journal of Biological Databases and Curation. 2016;2016.

[17] Doughty E, Kertesz-Farkas A, Bodenreider O, Thompson G, Adadey A, Peterson T, et al. Toward an automatic method for extracting cancer- and other disease-related point mutations from the biomedical literature. Bioinformatics. 2010;27:408-15.

[18] Peng Y, Rios A, Kavuluru R, Lu Z. Extracting chemical–protein relations with ensembles of SVM and deep learning models. Database: The Journal of Biological Databases and Curation. 2018;2018.

[19] Weber L, Sänger M, Garda S, Barth F, Alt C, Leser U. Humboldt@ DrugProt: Chemical-Protein Relation Extraction with Pretrained Transformers and Entity Descriptions. Proceedings of the seventh BioCreative challenge evaluation workshop. 2021.

[20] Thorn CF, Klein TE, Altman RB. PharmGKB: the pharmacogenomics knowledge base. Pharmacogenomics: Springer; 2013. p. 311-20.

[21] Piñero J, Ramírez-Anguita JM, Saüch-Pitarch J, Ronzano F, Centeno E, Sanz F, et al. The DisGeNET knowledge platform for disease genomics: 2019 update. J Nucleic acids research. 2020;48:D845-D55.

[22] Wu Y, Luo R, Leung H, Ting H-F, Lam T-W. Renet: A deep learning approach for extracting gene-disease associations from literature. International Conference on Research in Computational Molecular Biology: Springer; 2019. p. 272-84.

[23] Lamurias A, Clarke LA, Couto FM. Extracting microRNA-gene relations from biomedical literature using distant supervision. PloS one. 2017;12:e0171929.

[24] Xu R, Wang Q. Automatic construction of a large-scale and accurate drug-side-effect association knowledge base from biomedical literature. Journal of Biomedical Informatics. 2014;51:191-9.

[25] Islamaj Doğan R, Kim S, Chatr-Aryamontri A, Wei C-H, Comeau DC, Antunes R, et al. Overview of the BioCreative VI Precision Medicine Track: mining protein interactions and mutations for precision medicine. Database: The Journal of Biological Databases and Curation. 2019;2019.

[26] Raj Kanakarajan K, Kundumani B, Sankarasubbu M. BioELECTRA: pretrained biomedical text encoder using discriminators. Proceedings of the 20th Workshop on Biomedical Language Processing2021. p. 143-54.

[27] Alrowili S, Vijay-Shanker K. BioM-transformers: building large biomedical language models with BERT, ALBERT and ELECTRA. Proceedings of the 20th Workshop on Biomedical Language Processing2021. p. 221-7.

[28] Gu Y, Tinn R, Cheng H, Lucas M, Usuyama N, Liu X, et al. Domain-specific language model pretraining for biomedical natural language processing. ACM Transactions on Computing for Healthcare. 2021;3:1-23.

[29] Zhou W, Huang K, Ma T, Huang J. Document-level relation extraction with adaptive thresholding and localized context pooling. Proceedings of the AAAI Conference on Artificial Intelligence2021. p. 14612-20.

[30] Wang J, Chen X, Zhang Y, Zhang Y, Wen J, Lin H, et al. Document-level biomedical relation extraction using graph convolutional network and multihead attention: algorithm development and validation. JMIR Medical Informatics. 2020;8:e17638.

[31] Airola A, Pyysalo S, Björne J, Pahikkala T, Ginter F, Salakoski T. All-paths graph kernel for protein-protein interaction extraction with evaluation of cross-corpus learning. BMC Bioinformatics. 2008;9:1-12.

[32] Zhang T, Leng J, Liu Y. Deep learning for drug–drug interaction extraction from the literature: a review. Briefings in Bioinformatics. 2019;21:1609-27.

[33] Luo L, Lai P-T, Wei C-H, Arighi CN, Lu Z. BioRED: A Rich Biomedical Relation Extraction Dataset. Briefings in Bioinformatics. 2022.

[34] Lin C, Miller TA, Dligach D, Sadeque F, Bethard S, Savova G. A BERT-based One-Pass Multi-Task Model for Clinical Temporal Relation Extraction. BioNLP2020.

[35] Yadav S, Ramesh S, Saha S, Ekbal A. Relation extraction from biomedical and clinical text: Unified

multitask learning framework. IEEE/ACM Transactions on Computational Biology Bioinformatics. 2020;19:1105-16.

[36] Peng Y, Chen Q, Lu Z. An empirical study of multi-task learning on BERT for biomedical text mining. 2020 Workshop on Biomedical Natural Language Processing (BioNLP 2020)2020.

[37] Zhao S, Liu T, Zhao S, Wang F. A neural multi-task learning framework to jointly model medical named entity recognition and normalization. Proceedings of the AAAI Conference on Artificial Intelligence2019. p. 817-24.

[38] Wiatrak M, Iso-Sipila J. Simple hierarchical multi-task neural end-to-end entity linking for biomedical text. Proceedings of the 11th International Workshop on Health Text Mining and Information Analysis2020. p. 12-7.

[39] Zhou B, Cai X, Zhang Y, Yuan X. An end-to-end progressive multi-task learning framework for medical named entity recognition and normalization. Proceedings of the 59th Annual Meeting of the Association for Computational Linguistics and the 11th International Joint Conference on Natural Language Processing (Volume 1: Long Papers)2021. p. 6214-24.

[40] Eberts M, Ulges AJapa. Span-based joint entity and relation extraction with transformer pre-training. 24th European Conference on Artificial Intelligence - ECAI 2020. 2019.

[41] Gurulingappa H, Rajput AM, Roberts A, Fluck J, Hofmann-Apitius M, Toldo L. Development of a benchmark corpus to support the automatic extraction of drug-related adverse effects from medical case reports. Journal of Biomedical Informatics. 2012;45:885-92.

[42] Krallinger M, Rabal O, Akhondi SA, Pérez MP, Santamaría J, Rodríguez GP, et al. Overview of the BioCreative VI chemical-protein interaction Track. Proceedings of the sixth BioCreative challenge evaluation workshop2017. p. 141-6.

[43] Tang H, Cao Y, Zhang Z, Cao J, Fang F, Wang S, et al. Hin: Hierarchical inference network for document-level relation extraction. Advances in Knowledge Discovery and Data Mining: 24th Pacific-Asia Conference, PAKDD 2020, Singapore, May 11–14, 2020, Proceedings, Part I 24: Springer; 2020. p. 197-209.

[44] Yao Y, Ye D, Li P, Han X, Lin Y, Liu Z, et al. DocRED: A Large-Scale Document-Level Relation Extraction Dataset. Proceedings of the 57th Annual Meeting of the Association for Computational Linguistics. Florence, Italy: Association for Computational Linguistics; 2019. p. 764-77.

[45] Piñero J, Ramírez-Anguita JM, Saüch-Pitarch J, Ronzano F, Centeno E, Sanz F, et al. The DisGeNET knowledge platform for disease genomics: 2019 update. Nucleic acids research. 2020;48:D845-D55.

[46] Tiktinsky A, Viswanathan V, Niezni D, Azagury DM, Shamay Y, Taub-Tabib H, et al. A Dataset for N-ary Relation Extraction of Drug Combinations. 2022 Conference of the North American Chapter of the Association for Computational Linguistics:
Human Language Technologies2022. p. 3190 - 203.

[47] Davis AP, Grondin CJ, Johnson RJ, Sciaky D, Wiegers J, Wiegers TC, et al. Comparative toxicogenomics database (CTD): update 2021. Nucleic acids research. 2021;49:D1138-D43.

[48] Wei C-H, Allot A, Leaman R, Lu Z. PubTator central: automated concept annotation for biomedical full text articles. Nucleic acids research. 2019;47:W587-W93.

[49] Lipscomb CE. Medical subject headings (MeSH). Bulletin of the Medical Library Association. 2000;88:265.

[50] Li Y, Chen Y, Qin Y, Hu Y, Huang R, Zheng Q. Protein-protein interaction relation extraction based on multigranularity semantic fusion. Journal of Biomedical Informatics. 2021;123:103931.

[51] Asada M, Miwa M, Sasaki Y. Integrating heterogeneous knowledge graphs into drug–drug interaction extraction from the literature. Bioinformatics. 2023;39:btac754.

[52] Tiktinsky A, Viswanathan V, Niezni D, Azagury DM, Shamay Y, Taub-Tabib H, et al. A Dataset for N-ary Relation Extraction of Drug Combinations. the 2022 Conference of the North American Chapter of the Association for Computational Linguistics: Human Language Technologies2022. p. 3190 - 203.

[53] Beltagy I, Lo K, Cohan A. SciBERT: A Pretrained Language Model for Scientific Text. Hong Kong, China: Association for Computational Linguistics; 2019. p. 3615-20.

## Supplementary Material

### Experiment A: The performance of BioREx using partial BioRED data

Because it is time-consuming and costly to create a multi-relation dataset, such as BioRED, the results of the previous experiments suggest that using our approach may effectively decrease the need for manual data annotation.

We conducted an experiment using different BioRED subsets to determine whether leveraging external datasets can achieve a similar (or better) performance compared to the original data. Specifically, we randomly sampled four subsets of the BioRED training data of different sizes for model development and evaluation on the independent BioRED test set. The detailed results are shown in Figure 1. With the eight external datasets, the performance of the models trained on five incremental sizes (100, 200, 300, 400, and 500 abstracts) of training data is improved significantly. In particular, the performance of the model trained on 60% of the training set (300 abstracts) achieved a better result (75.8%) than did the model trained on the entire training set of the original BioRED (74.4%). These results demonstrate that our approach can take advantage of external datasets to build reliable training data with fewer annotation efforts.

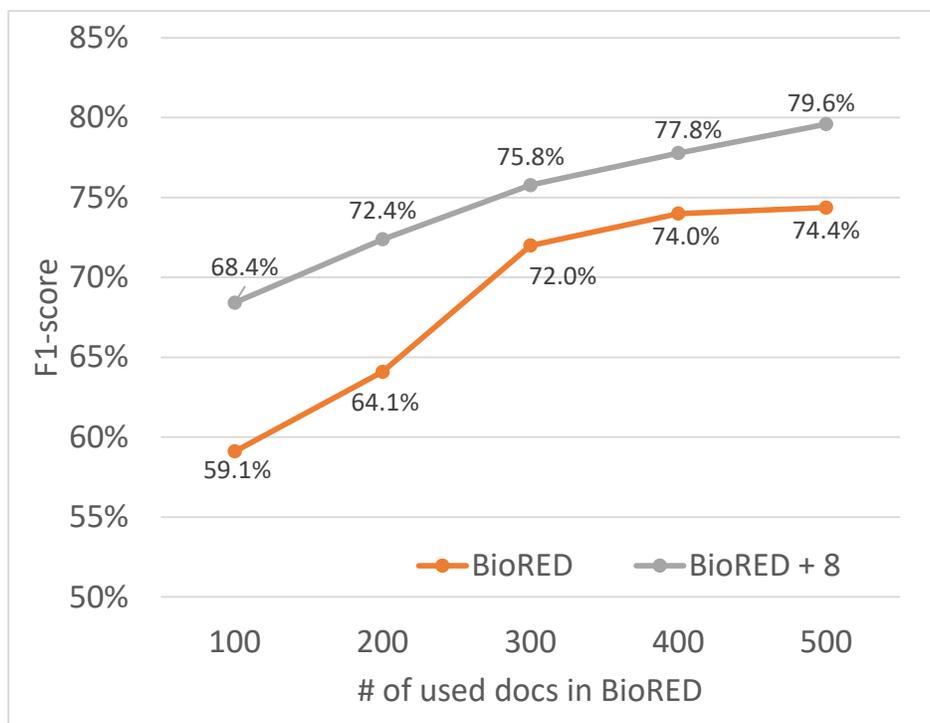

**Figure 1.** Performance of BioREx using partial BioRED with additional eight datasets.

We further compared the performance and efficiency of different PLM models. We chose five well-known pre-trained models for the comparison, including PubMedBERT [1], BioELECTRA [2], Bioformer [3], BioBert [4] and Roberta [5]. As shown in Table 1, we evaluated the final performance using the merged training set of BioRED with the eight corpora. PubMedBERT achieved the best performance, but Bioformer presents much more efficiency than do other models with close performance, as it is about two times faster on training and processing steps. Even though PubMedBERT achieved slightly higher performance than did Bioformer, Bioformer is more efficient than is PubMedBERT, which is an impressive advantage for processing large-scale data (e.g., entire PubMed abstracts, PMC full-text versions).

**Table 1.** Comparison of the PLM models

| PLM | Precision | Recall | F-score | Training time (per epoch) | Processing time per 1000 pairs | |
|---|---|---|---|---|---|---|
| | | | | | GPU | CPU |
| PubMedBERT [1] | 80.0 | **79.2** | **79.6** | 58m | 100s | 5.9m |
| BioELECTRA [2] | 81.2 | 76.4 | 78.7 | 1h4m | 90s | 5.4m |
| Bioformer [3] | 80.3 | 78.1 | 79.2 | 35m | 55s | 3.2m |
| BioBert [4] | **81.6** | 69.1 | 74.8 | 1h4m | 90s | 5.9m |
| Roberta [5] | 77.6 | 70.9 | 74.1 | 1h2m | 87s | 5.7m |

**Table S1.** Existing relation corpora and the available relations. The event detection corpora (e.g., BioNLP share task) are not listed below. G: Gene/Protein/Variant, C: Chemical/Drug, and D: Disease/Phenotype.

| Dataset | # Doc./Sent. | SEN/DOC | Relation | | | | | Not fully annotated (¥) Specific scope (†) Span is not provided (‡) Not available (×) |
|---|---|---|---|---|---|---|---|---|
| | | | <G,G> | <G,C> | <C,C> | <G,D> | <C,D> | |
| BioRED [6] | 600 abstracts | Document | ● | ● | ● | ● | ● | |
| AIMed[7] | 230 abstracts | Sentence | ● | | | | | |
| HPRD50[8] | 50 abstracts | Sentence | ● | | | | | |
| DrugProt [9] & ChemProt [10] | 5,000 abstracts | Sentence | | ● | | | | |
| DDI [11] | 905 abstracts | Sentence | | | ● | | | |
| RENET2 [12] | 500 full texts | Document | ● | | | | | ¥ Not fully annotated |
| RENET [13] | 30,192 abstracts | Document | ● | | | | | ¥ Not fully annotated |
| EU-ADR [14] | 300 abstracts | Sentence | | | ● | | ● | ● | ¥ Not fully annotated |
| N-ary dataset[15] | 1,634 abstracts | Sentence | | ● | | | | † Drug combination |
| ADE [16] | 2,972 abstracts | Sentence | | | | | ● | † Dose relation |
| BRONCO [17] | 108 full texts | Document | | | | ● | ● | † Genetic relation |
| BC5CDR[18] | 1,500 abstracts | Document | | | | | ● | † Induce relation |
| BioCreative VI PM[19] | 5,509 abstracts | Document | ● | | | | | † Genetic relation |
| PGxCorpus [20] | 945 sentences | Sentence | | ● | | ● | ● | † Phenotype relation |
| EMU [21] | 110 abstracts | Document | | | | ● | | † Variant relation ‡ Entity spans are not provided |
| PharmGKB [22] | - | - | | ● | | | | ‡ Repository |
| DisGeNet [23] | - | - | ● | | | | | ‡ Repository |
| CTD [24] | - | - | | ● | | ● | ● | ‡ Repository |
| GWAS [25] | - | - | | | | ● | | ‡ Repository |
| BindingDB [26] | - | - | | | ● | | | ‡ Repository |
| BioInfer[27] | 1,100 sentences | Sentence | ● | | | | | × |
| LLL[28] | 167 sentences | Sentence | ● | | | | | × |

| IEPA[29] | 300 abstracts | Document | | | ● | | × |
|---|---|---|---|---|---|---|---|
| BioCreative II PPI IPS [30] | 1,098 full texts | Document | ● | | | | × |
| BioCreative II.5 IPT [31] | 122 full texts | Document | ● | | | | × |
| n2c2 2018 ADE[32] | 505 summaries | - | ● | | | | × |

**References**

1. Gu, Y., et al., *Domain-specific language model pretraining for biomedical natural language processing.* ACM Transactions on Computing for Healthcare, 2021. **3**(1): p. 1-23.
2. Raj Kanakarajan, K., B. Kundumani, and M. Sankarasubbu. *BioELECTRA: pretrained biomedical text encoder using discriminators.* in *Proceedings of the 20th Workshop on Biomedical Language Processing.* 2021.
3. Fang, L. and K. Wang. *Team Bioformer at BioCreative VII LitCovid Track: Multic-label topic classification for COVID-19 literature with a compact BERT model.* in *Proceedings of the seventh BioCreative challenge evaluation workshop.* 2021.
4. Lee, J., et al., *BioBERT: a pre-trained biomedical language representation model for biomedical text mining.* Bioinformatics, 2020. **36**(4): p. 1234-1240.
5. Lewis, P., et al. *Pretrained language models for biomedical and clinical tasks: Understanding and extending the state-of-the-art.* in *Proceedings of the 3rd Clinical Natural Language Processing Workshop.* 2020.
6. Luo, L., et al., *BioRED: A Rich Biomedical Relation Extraction Dataset.* (in press) Briefings in Bioinformatics, 2022.
7. Bunescu, R., et al., *Comparative experiments on learning information extractors for proteins and their interactions.* Artificial Intelligence in Medicine, 2005. **33**(2): p. 139-155.
8. Fundel, K., R. Küffner, and R. Zimmer, *RelEx—Relation extraction using dependency parse trees.* Bioinformatics, 2007. **23**(3): p. 365-371.
9. Miranda, A., et al. *Overview of DrugProt BioCreative VII track: quality evaluation and large scale text mining of drug-gene/protein relations.* in *Proceedings of the seventh BioCreative challenge evaluation workshop.* 2021.
10. Krallinger, M., et al. *Overview of the BioCreative VI chemical-protein interaction Track.* in *Proceedings of the sixth BioCreative challenge evaluation workshop.* 2017.
11. Herrero-Zazo, M., et al., *The DDI corpus: An annotated corpus with pharmacological substances and drug–drug interactions.* Journal of Biomedical Informatics, 2013. **46**(5): p. 914-920.
12. Su, J., et al., *RENET2: high-performance full-text gene–disease relation extraction with iterative training data expansion.* NAR Genomics Bioinformatics, 2021. **3**(3): p. lqab062.
13. Wu, Y., et al. *Renet: A deep learning approach for extracting gene-disease associations from literature.* in *International Conference on Research in Computational Molecular Biology.* 2019. Springer.
14. Van Mulligen, E.M., et al., *The EU-ADR corpus: annotated drugs, diseases, targets, and their relationships.* Journal of Biomedical Informatics, 2012. **45**(5): p. 879-884.
15. Peng, N., et al., *Cross-sentence n-ary relation extraction with graph lstms.* Transactions of the Association for Computational Linguistics, 2017. **5**: p. 101-115.
16. Gurulingappa, H., et al., *Development of a benchmark corpus to support the automatic extraction of drug-related adverse effects from medical case reports.* Journal of Biomedical Informatics, 2012. **45**(5): p. 885-892.
17. Lee, K., et al., *BRONCO: Biomedical entity Relation ONcology COrpus for extracting gene-variant-disease-drug relations.* Database: The Journal of Biological Databases and Curation, 2016. **2016**.
18. Wei, C.-H., et al., *Assessing the state of the art in biomedical relation extraction: overview of the BioCreative V chemical-disease relation (CDR) task.* Database: The Journal of Biological Databases and Curation, 2016. **2016**.

19.    Islamaj Doğan, R., et al., *Overview of the BioCreative VI Precision Medicine Track: mining protein interactions and mutations for precision medicine.* Database: The Journal of Biological Databases and Curation, 2019. **2019**.

20.    Legrand, J., et al., *PGxCorpus, a manually annotated corpus for pharmacogenomics.* Scientific data, 2020. **7**(1): p. 1-13.

21.    Doughty, E., et al., *Toward an automatic method for extracting cancer- and other disease-related point mutations from the biomedical literature.* Bioinformatics, 2010. **27**(3): p. 408-415.

22.    Thorn, C.F., T.E. Klein, and R.B. Altman, *PharmGKB: the pharmacogenomics knowledge base*, in *Pharmacogenomics*. 2013, Springer. p. 311-320.

23.    Piñero, J., et al., *The DisGeNET knowledge platform for disease genomics: 2019 update.* J Nucleic acids research, 2020. **48**(D1): p. D845-D855.

24.    Davis, A.P., et al., *Comparative toxicogenomics database (CTD): update 2021.* J Nucleic acids research, 2021. **49**(D1): p. D1138-D1143.

25.    Johnson, A.D. and C.J. O'Donnell, *An Open Access Database of Genome-wide Association Results.* BMC Medical Genetics, 2009. **10**(1): p. 1-17.

26.    Gilson, M.K., et al., *BindingDB in 2015: a public database for medicinal chemistry, computational chemistry and systems pharmacology.* Nucleic acids research, 2016. **44**(D1): p. D1045-D1053.

27.    Pyysalo, S., et al., *BioInfer: a corpus for information extraction in the biomedical domain.* BMC Bioinformatics, 2007. **8**(1): p. 1-24.

28.    Nédellec, C. *Learning language in logic-genic interaction extraction challenge*. in *4. Learning language in logic workshop (LLL05)*. 2005. ACM-Association for Computing Machinery.

29.    Ding, J., et al., *Mining MEDLINE: abstracts, sentences, or phrases?*, in *Biocomputing 2002*. 2001, World Scientific. p. 326-337.

30.    Krallinger, M., et al., *Overview of the protein-protein interaction annotation extraction task of BioCreative II.* Genome biology, 2008. **9**(2): p. 1-19.

31.    Leitner, F., et al., *An Overview of BioCreative II.5.* IEEE/ACM Transactions on Computational Biology and Bioinformatics, 2010. **7**(3): p. 385-399.

32.    Henry, S., et al., *2018 n2c2 shared task on adverse drug events and medication extraction in electronic health records.* Journal of the American Medical Informatics Association, 2019. **27**(1): p. 3-12.